\documentclass{article}

\usepackage[preprint]{neurips_2025}
\usepackage{graphicx}
\usepackage{wrapfig}
\usepackage{lipsum} 
\usepackage{siunitx} 
\usepackage{amsmath}    
\usepackage[utf8]{inputenc} 
\usepackage[T1]{fontenc}    
\usepackage{hyperref}       
\usepackage{url}            
\usepackage{booktabs}       
\usepackage{amsfonts}       
\usepackage{nicefrac}       
\usepackage{microtype}      
\usepackage{xcolor}         
\usepackage{neurips_2025}
\usepackage{graphicx}    
\graphicspath{{figures/}}    
\usepackage{multirow}
\usepackage{subcaption}  
\usepackage{makecell}
\usepackage{ulem}
\usepackage{amssymb}

\title{Towards Text-free Graph Foundation Models: Rethinking Multi-Domain Graph Contrastive Learning}

\author{%
  ZihaoZhao, 
  Xinlong Zhai, 
  Jinyu Yang, 
  Chuan Shi\\
  Beijing University of Posts and Telecommunications, Beijing, China \\
  \small \{zihaozhao, zhaijojo, jinyu.yang, shichuan\}@bupt.edu.cn
}

\begin{document}

\maketitle

\begin{abstract}
Foundation models have achieved great success in natural language processing (NLP) and computer vision (CV). Their success largely stems from the ability to integrate multi-domain knowledge in pre-training and transfer it to target domains. Considering graph data, especially graphs without textual features, is ubiquitous in real-world applications such as social networks and recommendation systems, some researchers have attempted to extend this paradigm to the graph field, aiming to construct graph foundation models. However, unlike CV and NLP, there are huge gaps among the semantics and properties of graphs in different domains, while current works still adopt traditional contrastive pre-training strategies designed in the single-domain scenario, which regard contrastive samples from different domains as equivalent. From experimental investigations, we discovered that inherent domain-specific differences prevent these strategies from effectively absorbing knowledge from different domains to generate informative representations. In this paper, we propose a novel multi-domain pre-training and cross-domain transfer framework, namely MDGCL.In the pre-training stage, we design a contrastive learning strategy to substantially recognize and capture domain differences, and introduce domain tokens to encode domain-level global information. In the downstream stage, we introduce a domain attention mechanism to enable fine-grained domain knowledge transfer. Extensive experiments on five benchmark datasets have demonstrated that our method outperforms state-of-the-art significantly, with the maximum improvement of 19.33\% on accuracy and 19.13\% on Macro-F1 score.
\end{abstract}

\section{Introduction}
\label{sec:1}

In recent years, foundation models have achieved significant success in NLP \cite{llama, NLP_LM_1, gpt4} and CV \cite{CV_LM_1, CV_LM_2, CV_LM_3, CV_LM_4}.
They are pre-trained on large-scale data from multiple domains, and can generalize to a wide range of target domains and downstream tasks. 
Graph-structured data, as an important form of non-Euclidean data, offers unique advantages in many real-world applications\cite{GNN_app_1, GNN_app_2, GNN_app_3}. 
Therefore, we aim to replicate the success of foundation models in the field of graph learning, with the goal of enhancing the generalization ability of graph models across different domains.
To this end, we need to achieve multi-domain pre-training and cross-domain knowledge transfer on graph-structured data.

However, most existing graph pre-training methods \cite{GCL, GraphPrompt, AllinOne, GPF, GPPT, SimGrace} have not yet effectively achieved this goal.
These methods are initially designed for cross-task knowledge transfer within a single domain, assuming that the pre-training and downstream graphs originate from the same domain.
As a result, they are not well-suited for multi-domain pre-training scenarios, where domain differences often lead to a severe drop in their performance.

To utilize knowledge from multiple domains and build a more generalizable model, recent studies\cite{HiGPT, OneforAll, LLM_GNN} have explored using textual features as a bridge to unify graphs from different domains, with Large Language Models (LLMs)\cite{llama, gpt4} being employed to extract multi-domain knowledge.
However, they are inherently restricted to text-attributed graphs\cite{TAG_1, TAG_2}, but cannot generalize to text-free graphs, which are more widely applicable to model various real-world association data.
Few recent works \cite{GCOPE, SAMGPT} have begun to explore multi-domain pre-training on text-free graphs. They attempt to extract knowledge from different domains through the alignment of features and structures.
Unfortunately, they still adopt traditional contrastive pre-training approaches designed under a single-domain assumption, where graphs from the same domain and different domains serve as equivalent negative pairs, thereby failing to recognize the differences among domains.

To address these limitations, we propose MDGCL: \textbf{M}ulti-\textbf{D}omain \textbf{G}raph \textbf{C}ontrastive \textbf{L}earning, 
a novel framework for multi-domain pre-training and cross-domain transfer.
During the pre-training stage, we treat two subgraphs from the same domain as positive pairs and from different domains as negative pairs.
By requiring the GNN to distinguish whether a pair of subgraphs belongs to the same domain, we improve its capacity to recognize domain-specific differences.
Simultaneously, we assign each source domain a domain token that encodes domain-specific prior knowledge and then insert these tokens into each pair of subgraphs to serve as connectors, thereby enabling the GNN model to capture domain-level global information.
In the downstream stage, we also introduce a domain-level attention mechanism.
Specifically, we compute the similarity between each downstream node and source domain tokens, utilizing the attention-based aggregation to enhance the node features.
Through this enhancement, we effectively capture the correlations between the target domain and source domains, thereby enhancing cross-domain knowledge transfer.

To the best of our knowledge, we are the first to propose the contrastive learning strategy specifically for multi-domain scenarios.
Elaborate experiments on five benchmark datasets show that our method outperforms state-of-the-art steadily and significantly, with the maximum improvement of 19.33\% on accuracy and 19.13\% on Macro-F1 score.
We also test our method on different numbers of source domains, which demonstrates our method has the potential for pre-training on large-scale data and serve as inchoate text-free graph foundation models.

\section{Investigation for Representations of Different Domains by Text-free Graph Foundation Models}
\label{sec:2}

Some works\cite{GCOPE,SAMGPT} have begun to explore text-free graph foundation models, but they still adopt conventional contrastive pre-training strategies, which are initially designed for single-domain scenarios.
The graph field differs significantly from the fields of computer vision (CV) and natural language processing (NLP), as each graph exhibits notable domain-specific differences.
For instance, Cora\cite{Cora_Citeseer} and CiteSeer\cite{Cora_Citeseer} are citation networks, where nodes represent academic papers, and edges denote citation relationships.
In contrast, Squirrel\cite{Squi_Chame} denotes a Wikipedia page-page network, where nodes represent articles about squirrels from the English Wikipedia, and edges reflect mutual links among them.
Apparently, Cora and CiteSeer are similar domains, whereas Squirrel exhibits a significant difference from them.

\begin{figure}
\centering
\begin{subfigure}[c]{0.12\textwidth}
  \includegraphics[width=\textwidth]{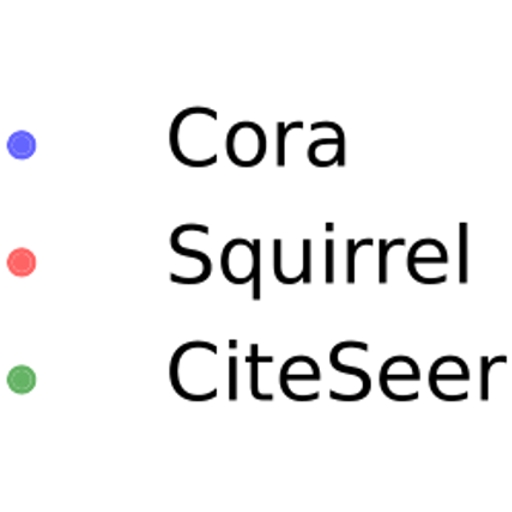}
  \label{fig:dist_label}
\end{subfigure}
\begin{subfigure}[c]{0.20\textwidth}
  \includegraphics[width=\textwidth]{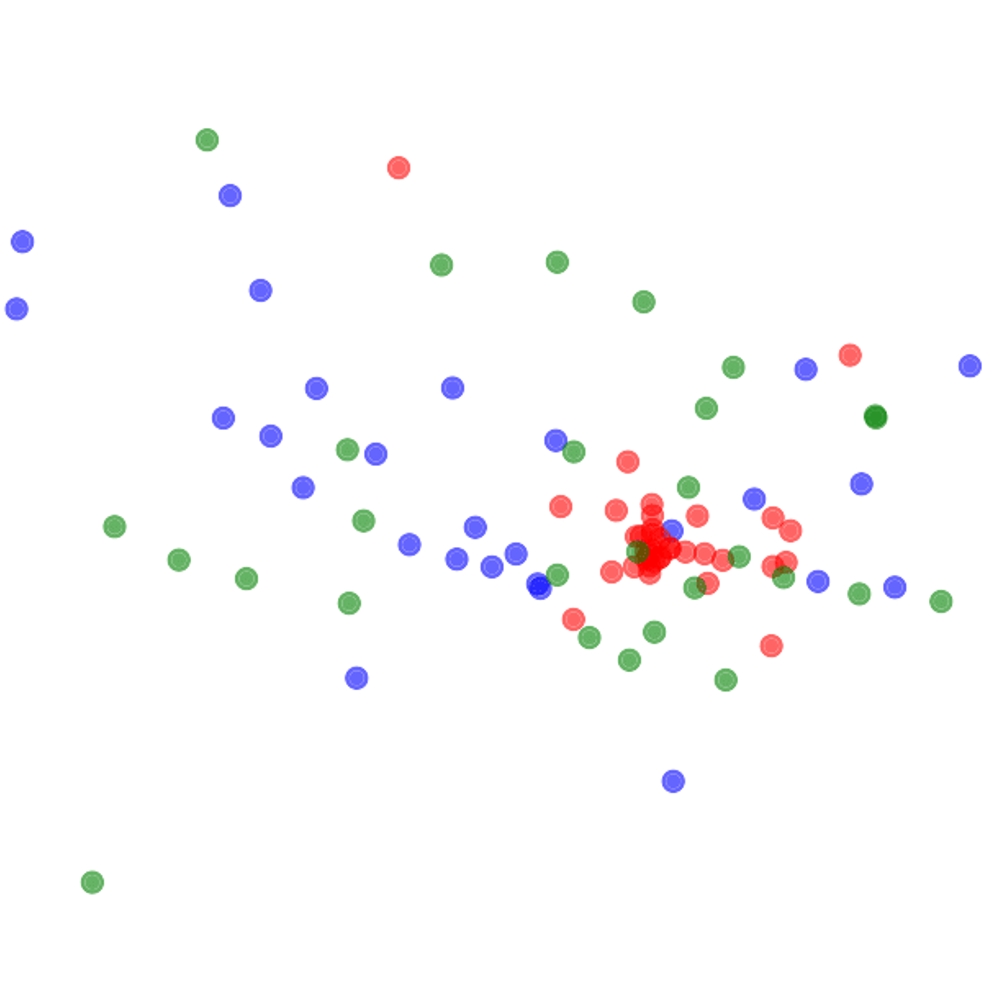}
  \caption{Initial features}
  \label{fig:dist_initial}
\end{subfigure}
\hfill
\begin{subfigure}[c]{0.20\textwidth}
  \includegraphics[width=\textwidth]{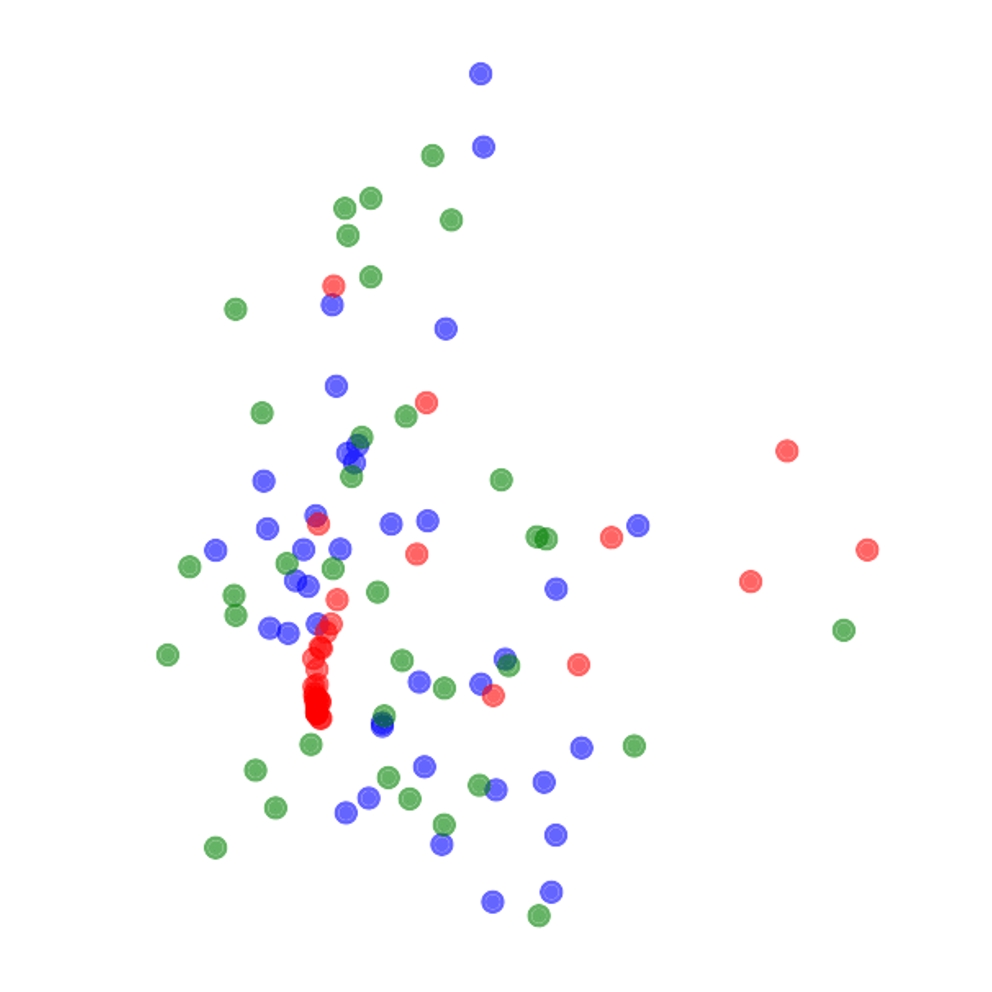}
  \caption{SAMGPT}
  \label{fig:dist_samgpt}
\end{subfigure}
\hfill
\begin{subfigure}[c]{0.20\textwidth}
  \includegraphics[width=\textwidth]{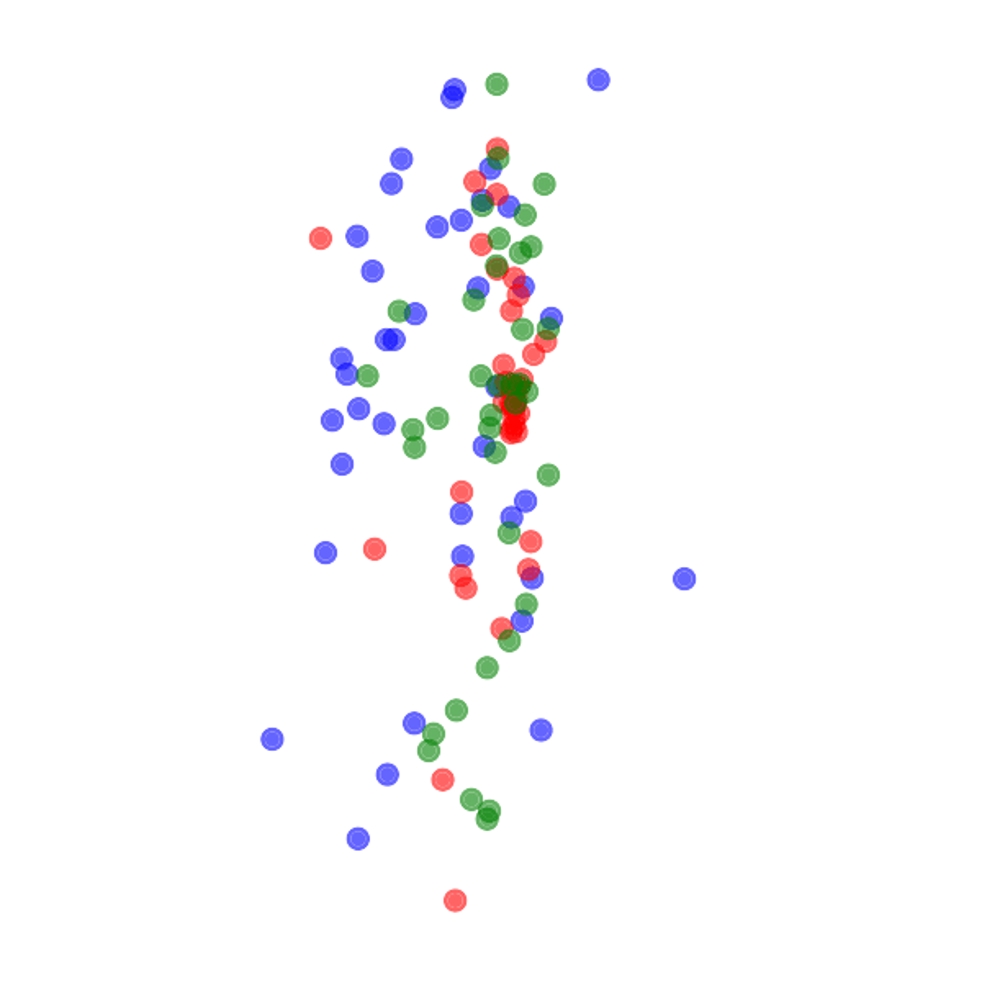}
  \caption{GCOPE}
  \label{fig:dist_gcope}
\end{subfigure}
\hfill
\begin{subfigure}[c]{0.20\textwidth}
  \includegraphics[width=\textwidth]{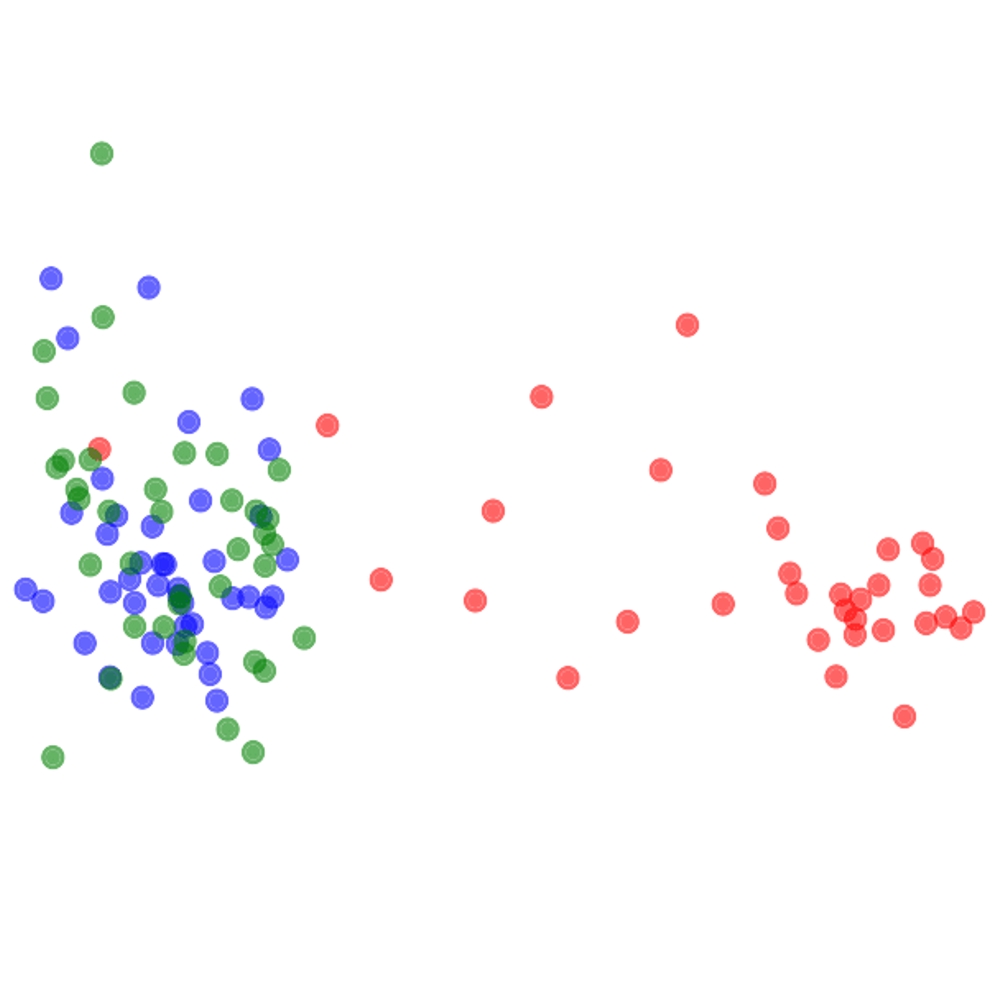}
  \caption{MDGCL}
  \label{fig:dist_mdgcl}
\end{subfigure}
\caption{Visualization for initial features and learned representations of SAMGPT, GCOPE and MDGCL.}
\label{fig:invest}
\end{figure}

Based on the aforementioned three datasets, we analyze the ability of three text-free graph foundation models: GCOPE, SAMGPT, and MDGCL, to capture domain-specific differences.
Specifically, we leverage these models to perform multi-domain pre-training on four source domain datasets: CiteSeer, PubMed\cite{Pubmed}, Photo\cite{Photo}, Computers\cite{computers}.
Then, we take Cora, CiteSeer, and Squirrel as test datasets and apply the pre-trained GNN models to generate node representations on them.
We randomly select approximately 40 nodes for every test dataset with balanced labels, illustrating the representations generated by the three methods, as well as the initial features in Figure \ref{fig:invest}.

Figure \ref{fig:invest}(b) and Figure \ref{fig:invest}(c) illustrate that SAMGPT and GCOPE map the node representations of all three domains to a similar distribution, proving that SAMGPT and GCOPE cannot capture domain-specific differences.
The reason is that they still adopt traditional pre-training strategies designed for the single-domain scenario, treating subgraphs from different domains and from the same domain as equivalent negative pairs, thereby neglecting domain differences.
More critically, when domain differences are substantial, these traditional pre-training methods often lead to negative transfer\cite{GCOPE}.

In contrast, our proposed MDGCL, which requires the model to determine whether two subgraphs belong to the same domain, adaptively maps nodes of similar domains to closer distributions, while mapping nodes of radically different domains to more distant distributions, as shown in Figure \ref{fig:invest}(d).
This demonstrates our method can more effectively capture domain differences and is better suited for multi-domain pre-training scenarios.

\section{Preliminaries}
\label{sec:3}

\paragraph{Problem definition}
In this paper, we focus on multi-domain pre-training and cross-domain transfer scenarios, where a GNN model $f$ will be pre-trained on graphs from different source domains and then adapted to a target domain unseen during pre-training.

In the pre-training stage, we are given a collection of graphs $\mathcal{G}_s=\left\{G_1, G_2, \ldots, G_M\right\}$.Each $G_i=\left(V_i, E_i, X_i\right)$ represents a source domain, where $V_i$ is the set of nodes , $E_i$ is the set of edges , and $X_i \in \mathbb{R}^{|V_i| \times d_i}$ is the corresponding feature matrix.It is important to note that each graph has a distinct initial feature dimension $d_i$.
Similarly, the downstream dataset is defined on a single graph $G_t=\left(V_t, E_t, X_t\right)$, where $t$ stands for the target domain and $G_t \notin \mathcal{G}_s$.

In downstream tasks, we focus on few-shot node and graph classification. 
For node classification, each node $v_i$ in the downstream graph  $G_t$ is assigned a label $y_i \in Y$, where $Y$ represents the set of node classes. 
For graph classification, following previous works \cite{GCOPE, SAMGPT}, we extract induced subgraphs around the labeled nodes in $G_t$, assigning each one the label of its central node.
And then we perform graph classification on these induced graphs.

\paragraph{Graph neural networks.}
Graph Neural Networks (GNNs)\cite{GCN, GraphSAGE, GAT, GraphTransformer} have emerged as the dominant approach to learn representations on graph-structured data. 
Most GNNs are based on the message-passing paradigm, where each node updates its representation by aggregating information from its neighbors over multiple iterations.
Formally, a GNN model can be abstracted as a function $f$ that maps a graph $G=\left(V, E, X\right)$ to node-level representations $H$.
\begin{equation}
    f:(V, E, X)\to H
\label{Eq:GNN}
\end{equation}
where $H\in\mathbb{R}^{|V|\times d}$ denotes the output node embeddings.
In graph-level tasks, a $ReadOut$ function is typically applied to the node embeddings $H$ to generate a graph-level representation.

\section{Method}
\label{headings}

\begin{figure}[htbp]
    \centering
    \includegraphics[width=\linewidth]{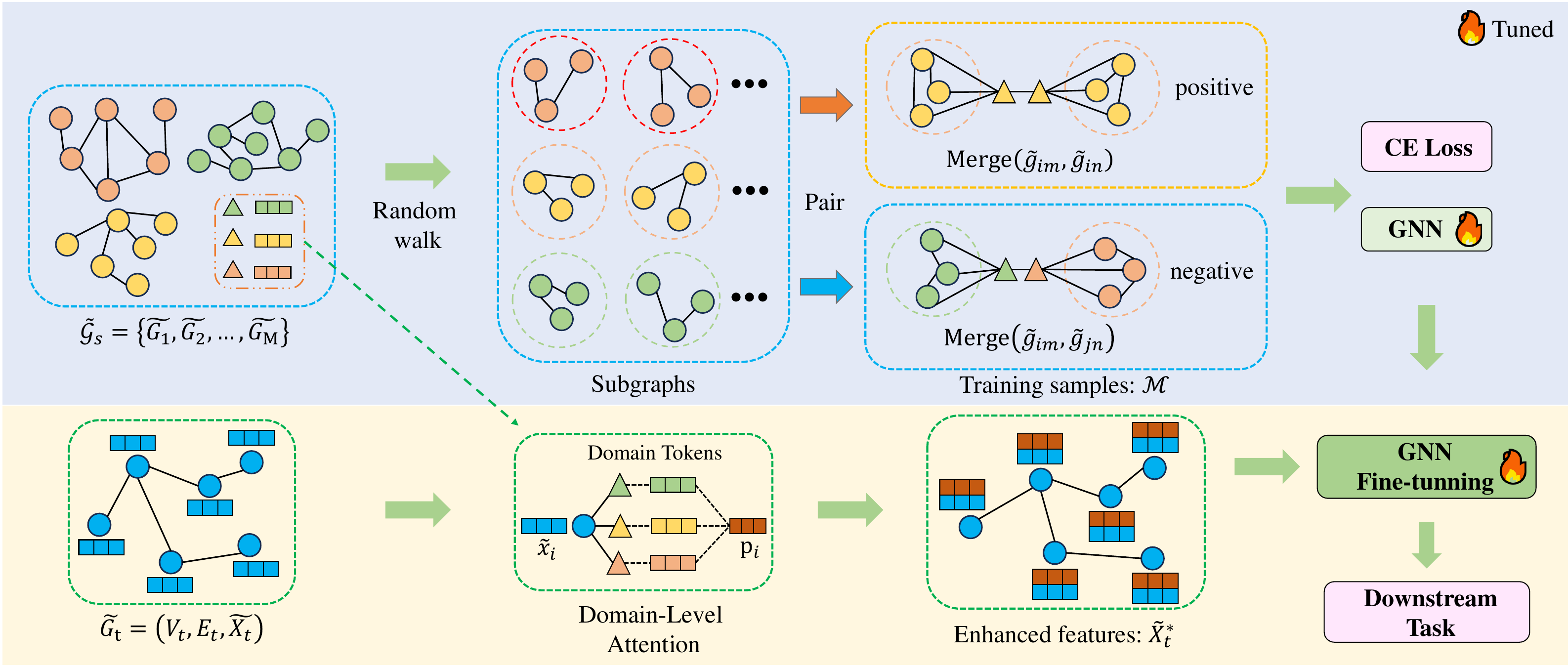}
    \caption{Overall framework of MDGCL. The upper part represents the pre-training stage, and the lower part denotes the downstream stage.}
    \label{FIG:MDGCL}
\end{figure}

\subsection{Overview}
In this section, we present a novel method for multi-domain pre-training and cross-domain transfer. 
The overall framework is shown in Figure \ref{FIG:MDGCL}.
In the pre-training stage, we design a new pre-training objective to recognize domain discrepancies.
Additionally, we introduce domain tokens to capture domain-specific global information. 
In the downstream stage, we propose the domain-level attention mechanism to facilitate knowledge transfer from source domains to the target domain. The time complexity analysis of our method can be found in the Appendix.

\subsection{Pre-training stage}
We design a novel pre-training objective that requires the model to determine whether a pair of subgraphs originates from the same domain, thereby promoting the recognition of domain-specific differences. 
In addition, we introduce domain tokens to connect each subgraph pair to a unified graph representation.
These tokens not only capture domain-specific global information but also facilitate the integration of knowledge across multiple domains.
The pre-training stage is illustrated in the upper part of Figure \ref{FIG:MDGCL}.

\subsubsection{Feature dimension unifying}

As mentioned in Section \ref{sec:3}, during the pre-training stage, we have a set of graphs $\mathcal{G}_s=\left\{G_1, G_2, \ldots, G_M\right\}$, where each graph $G_i=\left(V_i, E_i, X_i\right)$ represents a source domain. 
Since the initial feature dimension of each graph is not consistent, we first employ a mapping function to unify them to a same dimension:

\begin{equation}
    \tilde{X}_i=\operatorname{Map}\left(X_i\right) \in \mathbb{R}^{\left|\mathcal{V}_i\right| \times \tilde{d}}
\label{Eq:eq_1}
\end{equation}

where $\operatorname{Map}$ denotes the mapping function, and $\tilde{d}$ is a hyperparameter that defines the unified feature dimension.
We follow previous works \cite{GCOPE,SAMGPT} and employ Singular Value Decomposition (SVD)\cite{SVD} to implement the mapping function.
Each dimension-unified source graph is denoted as $\tilde{G}_i=\left(V_i, E_i, \tilde{X}_i\right)$.

\subsubsection{Generating training samples}
\label{sec:Merge}

With dimension-unified source graphs $\tilde{\mathcal{G}_s}=\left\{\tilde{G}_1, \tilde{G}_2, \ldots, \tilde{G}_M\right\}$, 
we employ random-walk based sampling\cite{PageRank} to generate $K$ subgraphs $\left\{\tilde{g}_{i1}, \tilde{g}_{i2}, \ldots, \tilde{g}_{iK}\right\}$ for each source graph $\tilde{G}_i$.
Then we leverage these subgraphs to construct positive and negative sample pairs.
Specifically, any pair of subgraphs $(\tilde{g}_{im}, \tilde{g}_{in})$ sampled from the same domain $\tilde{G}_i$ is regarded as a positive pair, 
while the pair of subgraphs sampled from different domains $(\tilde{g}_{im}, \tilde{g}_{jn})$, where $i\neq j$, constitutes a negative pair.

Additionally, we introduce domain tokens to help GNN encode domain-level global information.
We assign a domain token $t_i$ for each source domain $\tilde{G}_i$. 
In this paper, we use the $\operatorname{sum}$ operation to construct domain tokens, which represent the domain-specific global information:
\begin{equation}t_{i}=\operatorname{sum}(\tilde{X}_{i})\end{equation}

For each subgraph pair $(\tilde{g}_{im}, \tilde{g}_{jn})$, we connect all nodes in both subgraphs to their corresponding domain tokens $t_i$ and $t_j$ respectively, with an additional connection between $t_i$ and $t_j$.
We denote this operation as $\operatorname{Merge}$ and the resulting graph as $ \operatorname{Merge}(\tilde{g}_{im}, \tilde{g}_{jn}) $.
The set of all graphs obtained through the \( \operatorname{Merge} \) operation and their labels is denoted as  \( \mathcal{M} = \{ (\operatorname{Merge}(\tilde{g}_{im}, \tilde{g}_{jn}), y_{i, j}) \mid i, j \in [1, M], m, n \in [1, K] \} \), 
where $y_{i, j} \in \{0, 1\}$ is a binary classification label indicating whether the subgraphs originate from the same domain.

We take the set \( \mathcal{M} \) as training samples and require the GNN model $f$ to distinguish whether the two subgraphs in each \( \operatorname{Merge}(\tilde{g}_{im}, \tilde{g}_{jn}) \) originate from the same domain, 
thereby enhancing the model's ability to recognize and capture domain-specific differences.

\subsubsection{Multi-domain pre-training}

We implement our proposed multi-domain pre-training objective as the following cross-entropy loss function:
\begin{equation}
\mathcal{L}_{\operatorname{pre}}(f, \operatorname{Proj}_{\operatorname{pre}}) = \operatorname{CrossEntropy}(\hat{y}_{ij}, y_{ij}) , \quad \hat{y}_{ij} = \operatorname{Proj}_{\operatorname{pre}}(\operatorname{ReadOut}(f(\operatorname{Merge}(\tilde{g}_{im}, \tilde{g}_{jn}))))
\label{Eq:Pretrain}
\end{equation}
where the $\operatorname{ReadOut}$ function aggregates the output node embeddings of GNN encoder $f$ into a graph-level representation.
And the \( \operatorname{Proj}_{\operatorname{pre}} \) denotes a projection head that maps the graph representation to a prediction vector $\hat{y}_{ij}$, predicting whether the two subgraphs come from the same domain.
We optimize this pre-training loss by updating the parameters of the GNN $f$ and the projection head \( \operatorname{Proj}_{\operatorname{pre}} \).

\subsection{Downstream stage}
In the downstream stage, we propose the domain-level attention mechanism that computes the similarity between the downstream node and each source domain token, aiming to capture the correlations between the target domain and source domains. Then we use the attentive aggregation as feature enhancement, enabling fine-grained, domain-level knowledge transfer.
The downstream stage is illustrated in the lower part of Figure \ref{FIG:MDGCL}.

\subsubsection{Domain-level attention mechanism}

As mentioned in Section \ref{sec:3}, we denote the downstream target domain as $G_t = (V_t, E_t, X_t)$. 
Similar to the pre-training stage, we first apply the mapping function to align the feature dimension of the downstream graph \( G_t \) with the unified dimension \( \tilde{d} \) :

\begin{equation}
    \tilde{X}_t=\operatorname{Map}\left(X_t\right) \in \mathbb{R}^{\left|V_t\right| \times \tilde{d}}
\label{Eq:eq_2}
\end{equation}
With dimension-aligned target graph $\tilde{G}_t=\left(V_t, E_t, \tilde{X}_t\right)$, 
we compute attention scores between each downstream node \( v_i \in V_t \) and all source domain tokens \( \mathcal{S} = \{ t_1, t_2, \dots, t_M \} \), 
which can capture correlations between the target domain and source domains.
Subsequently, we utilize the attentive aggregation of source domain tokens to perform domain-level knowledge transfer:

\begin{equation}
p_i = \sum_{m=1}^{M} \alpha_{im} \cdot W_v t_m , \quad \alpha_{im} = \frac{\exp(W_k t_m \cdot W_q \tilde{x}_i)}{\sum_{l=1}^{M} \exp(W_k t_l \cdot W_q \tilde{x}_i)}
\end{equation}

where \( \tilde{x}_i \in \tilde{X}_t \) is the feature vector of downstream node \( v_i \) after dimension alignment , 
\( p_i \) represents the domain-level enhancement of node \( v_i \), 
and \( \Theta = (W_q, W_k, W_v) \) denote the learnable parameters of the attention mechanism.
For each downstream node \( v_i \), we add \( p_i \) to its feature \( \tilde{x}_i \) to achieve knowledge transfer from source domains to the target domain.
It can be denoted as:

\begin{equation}
\tilde{X}_t = \{ \tilde{x}_1, \tilde{x}_2, \ldots, \tilde{x}_N \} \quad \tilde{X}_t^* = \{ \tilde{x}_1 + p_1, \tilde{x}_2 + p_2, \ldots, \tilde{x}_N + p_N \}
\end{equation}

The enhanced feature matrix $\tilde{X}_t^*$ replaces the initial features $\tilde{X}_t$ and will be utilized for downstream fine-tuning.

\subsubsection{Downstream fine-tuning}
For downstream node classification and graph classification, we define the training set
$ \mathcal{O} = \{(x_1, y_1), (x_2, y_2), \dots, (x_n, y_n)\} $, 
where each $ x_i$  corresponds to either a node or a graph instance, and $ y_i \in Y $ denotes its class label from the set of all possible labels $ Y $.
Then we optimize the fine-tuning loss:

\begin{equation}
\mathcal{L}_{\text{FT}}(f, \operatorname{Proj}_{\text{FT}}, \Theta) = \frac{1}{n} \sum_{i=1}^{n} \text{CrossEntropy}(\operatorname{Proj}_{\text{FT}}(h_{x_i}), y_i), \quad H_t = f\left(V_t, E_t, \tilde{X}_t^*\right)
\end{equation}

where $h_{x_i}$ represents the output embedding of the node or graph $x_i$ based on $H_t$. When \( x_i \) is a graph, we use a sum pooling $\operatorname{ReadOut}$ on \( H_t \) to obtain \( h_{x_i} \).
Additionally, \( f \) denotes the GNN encoder initialized with the parameters obtained from pre-training, and is tuned in the downstream stage to adapt to the target domain.
\( \operatorname{Proj}_{\text{FT}} \) refers to a task-specific projection head used for downstream prediction.

\section{Experiments}
\label{Experiment}
To evaluate the effectiveness of MDGCL, we perform extensive experiments on five benchmark datasets to answer the following questions:
\begin{itemize}

\item \textbf{RQ1:}
How does our proposed MDGCL method perform compared with baseline methods?
\item \textbf{RQ2:}
Is MDGCL universally compatible with homophilic and heterophilic graphs?
\item \textbf{RQ3:}
How does each component contribute to the overall performance of MDGCL?
\item \textbf{RQ4:}
As a graph foundation model, can the performance of MDGCL increase with the growth of the pre-training domain number?

\end{itemize}

Moreover, we claim our method has superiority with regard to the pre-training time cost compared with other text-free graph foundation model baselines, which is elaborated in the Appendix. The hyperparameter study can also be found in the Appendix.

\subsection{Experimental setup}
\label{sec:Experimental Setup}
\paragraph{Datasets.}
We conduct experiments on five widely used graph benchmark datasets.
Cora\cite{Cora_Citeseer}, CiteSeer\cite{Cora_Citeseer}, and PubMed\cite{Pubmed} are citation networks, 
where nodes correspond to academic documents and edges denote citation relationships.
Photo\cite{Photo} and Computers\cite{computers} are networks illustrating co-purchase relationships from Amazon, where each node is characterized by a bag-of-words representation of product reviews, and edges indicate frequent co-purchases between products.
More details about datasets are presented in the Appendix.

\paragraph{Baselines.}
We compare the performance of MDGCL against eight state-of-the-art methods, which can be categorized into three groups, as follows. 

(1) End-to-end graph neural networks. GCN\cite{GCN} and GAT\cite{GAT} are representative supervised GNN methods that aggregate information from local neighborhoods to update node representations. They are trained from scratch on each downstream task without any pre-training.

(2) Single-domain graph pre-training methods. GraphCL \cite{GCL} and SimGrace \cite{SimGrace} adopt a "pre-train and fine-tune" paradigm. GraphCL leverages augmentation-based contrastive learning to capture inherent graph properties, while SimGrace generates contrastive views by injecting perturbations into GNN parameters. Both of them tune a task-specific projection head during the downstream stage for prediction.
GraphPrompt\cite{GraphPrompt} and GPF\cite{GPF} adopt a "pre-train and prompt" paradigm. GraphPrompt employs a universal task template and a learnable prompt to bridge the gap between pre-training and downstream tasks. GPF incorporates a prompt token into the downstream input space to enhance predictive performance.

(3) Multi-domain graph pre-training methods. GCOPE\cite{GCOPE} and SAMGPT\cite{SAMGPT} are recent works that explore multi-domain pre-training for text-free graphs.
They attempt to pre-train a universal GNN model on graphs across different domains by alignment of feature and structure.
GCOPE aligns structural and semantic patterns through coordinators, while SAMGPT inserts structural tokens into the GNN architecture to facilitate structural alignment and transfer.

\paragraph{Setup of pre-training and downstream adaption.}
Following previous works\cite{GCOPE, SAMGPT} , we treat each dataset as an independent domain. 
We adopt a "multi-source to single-target" pipeline, designating each dataset as the target domain in turn, while leveraging the remaining four as source domains for pre-training.
To ensure fair comparison, we use SVD to unify the feature dimension to 50 and merge all source graphs for joint pre-training.
Notably, the single-domain approaches process all source graphs in an isolated manner, amalgamating all datasets into a batch without connections.

All pre-training methods except GCOPE adopt a two-layer GCN with a hidden dimension of 256, using the Adam optimizer with a learning rate of 0.0001.
For GCOPE, we employ a 2-layer FAGCN\cite{FAGCN} as the backbone for pre-training due to its superior performance over GCN in this framework.
In terms of MDGCL, we sample 50 subgraphs from each source domain during the pre-training stage.
Each domain token is computed via sum pooling over all node features of the corresponding source domain.
During the downstream stage, we adopt a multi-head attention with 2 heads as the domain-level attention mechanism.
More implemental details are listed in the Appendix.

For each target domain, we adopt node classification and graph classification as downstream tasks.
Following previous works\cite{GraphPrompt, GCOPE}, we apply the m-shot learning setting for each target dataset to build the training data and then split the rest randomly in 1:9 for validation and test.

{\large
\renewcommand{\arraystretch}{1.5}
\begin{table}[t]
\centering
\caption{One-shot graph classification performance (mean±std Acc/F1).
The best method in each column is bolded, and the runner-up is underlined.
}
\label{tab:Graph_2_metric}
\resizebox{1.0\textwidth}{!}{
\begin{tabular}{c cc cc cc cc cc}
\toprule
\multirow{2}{*}{Method} & \multicolumn{2}{c}{Cora} & \multicolumn{2}{c}{CiteSeer} & \multicolumn{2}{c}{PubMed} & \multicolumn{2}{c}{Photo} & \multicolumn{2}{c}{Computers} \\
\cmidrule(r){2-3} \cmidrule(r){4-5} \cmidrule(r){6-7} \cmidrule(r){8-9} \cmidrule(r){10-11}
& Acc & F1 & Acc & F1 & Acc & F1 & Acc & F1 & Acc & F1 \\
\midrule
GCN & 0.3133 \textsubscript{±.02} & 0.2774 \textsubscript{±.01} & 0.4219 \textsubscript{±.03} & 0.3677 \textsubscript{±.07} & 0.4114 \textsubscript{±.05} & 0.3233 \textsubscript{±.09} & 0.4553 \textsubscript{±.02} & 0.4697 \textsubscript{±.02} & 0.2721 \textsubscript{±.10} & 0.2218 \textsubscript{±.02} \\
GAT & 0.3513 \textsubscript{±.08} & 0.3133 \textsubscript{±.02} & 0.3715 \textsubscript{±.03} & 0.3081 \textsubscript{±.02} & 0.4059 \textsubscript{±.03} & 0.3714 \textsubscript{±.01} & 0.4719 \textsubscript{±.03} & 0.4342 \textsubscript{±.01} & 0.3283 \textsubscript{±.10} & 0.2517 \textsubscript{±.07} \\
\midrule
GraphCL & 0.2495 \textsubscript{±.04} & 0.2174 \textsubscript{±.04} & 0.3823 \textsubscript{±.04} & 0.3409 \textsubscript{±.04} & 0.4634 \textsubscript{±.08} & 0.4157 \textsubscript{±.11} & 0.4754 \textsubscript{±.06} & 0.4792 \textsubscript{±.08} & 0.3411 \textsubscript{±.11} & 0.2956 \textsubscript{±.09} \\
SimGRACE & 0.2307 \textsubscript{±.03} & 0.1828 \textsubscript{±.04} & 0.3669 \textsubscript{±.10} & 0.3343 \textsubscript{±.08} & \uline{0.5033} \textsubscript{±.08} & 0.4346 \textsubscript{±.11} & 0.3725 \textsubscript{±.09} & 0.3835 \textsubscript{±.06} & 0.2527 \textsubscript{±.01} & 0.2411 \textsubscript{±.09} \\
GPF & 0.3189 \textsubscript{±.09} & 0.2195 \textsubscript{±.07} & 0.4341 \textsubscript{±.08} & 0.3882 \textsubscript{±.07} & 0.4076 \textsubscript{±.08} & 0.2792 \textsubscript{±.09} & 0.4032 \textsubscript{±.07} & 0.4032 \textsubscript{±.06} & 0.3205 \textsubscript{±.13} & 0.2769 \textsubscript{±.10} \\
GraphPrompt & 0.3167 \textsubscript{±.03} & 0.2990 \textsubscript{±.02} & 0.3828 \textsubscript{±.09} & 0.3365 \textsubscript{±.09} & 0.3994 \textsubscript{±.08} & 0.2858 \textsubscript{±.07} & 0.5025 \textsubscript{±.08} & 0.4965 \textsubscript{±.06} & 0.4183 \textsubscript{±.08} & 0.3755 \textsubscript{±.06} \\
\midrule

GCOPE & 0.4130 \textsubscript{±.02} & \uline{0.4568} \textsubscript{±.01} & \uline{0.6161} \textsubscript{±.05} & \uline{0.5081} \textsubscript{±.05} & 0.4667 \textsubscript{±.01} & 0.4278 \textsubscript{±.01} & \uline{0.6487} \textsubscript{±.06} & 0.5336 \textsubscript{±.08} & 0.4213 \textsubscript{±.03} & 0.3279 \textsubscript{±.02} \\
SAMGPT & \uline{0.5428} \textsubscript{±.13} & 0.4115 \textsubscript{±.11} & 0.4229 \textsubscript{±.09} & 0.3721 \textsubscript{±.08} & 0.4942 \textsubscript{±.09} & \uline{0.4765} \textsubscript{±.12} & 0.5960 \textsubscript{±.11} & \uline{0.5863} \textsubscript{±.10} & \uline{0.4588} \textsubscript{±.11} & \textbf{0.4227} \textsubscript{±.09} \\
MDGCL & \textbf{0.5514} \textsubscript{±.02} & \textbf{0.5274} \textsubscript{±.02} & \textbf{0.6444} \textsubscript{±.02} & \textbf{0.6053} \textsubscript{±.03} & \textbf{0.5608} \textsubscript{±.02} & \textbf{0.5452} \textsubscript{±.03} & \textbf{0.6617} \textsubscript{±.05} & \textbf{0.6221} \textsubscript{±.02} & \textbf{0.4619} \textsubscript{±.04} & \uline{0.3985} \textsubscript{±.02} \\
\bottomrule
\end{tabular}
}
\end{table}
}

{\large
\renewcommand{\arraystretch}{1.5}
\begin{table}[t]
\centering
\caption{One-shot node classification performance (mean±std Acc/F1).
The best method in each column is bolded, and the runner-up is underlined.}
\label{tab:Node_2_metric}
\resizebox{1.0\textwidth}{!}{
\begin{tabular}{c cc cc cc cc cc}
\toprule
\multirow{2}{*}{Method} & \multicolumn{2}{c}{Cora} & \multicolumn{2}{c}{CiteSeer} & \multicolumn{2}{c}{PubMed} & \multicolumn{2}{c}{Photo} & \multicolumn{2}{c}{Computers} \\
\cmidrule(r){2-3} \cmidrule(r){4-5} \cmidrule(r){6-7} \cmidrule(r){8-9} \cmidrule(r){10-11}
& Acc & F1 & Acc & F1 & Acc & F1 & Acc & F1 & Acc & F1 \\
\midrule
GCN & 0.3055 \textsubscript{±.01} & 0.2960 \textsubscript{±.12} & 0.2719 \textsubscript{±.03} & 0.2335 \textsubscript{±.01} & 0.3517 \textsubscript{±.05} & 0.3746 \textsubscript{±.09} & 0.2996 \textsubscript{±.06} & 0.2794 \textsubscript{±.06} & 0.2984 \textsubscript{±.02} & 0.3175 \textsubscript{±.05} \\
GAT & 0.2717 \textsubscript{±.02} & 0.3015 \textsubscript{±.02} & 0.2239 \textsubscript{±.01} & 0.2107 \textsubscript{±.04} & 0.2975 \textsubscript{±.04} & 0.3059 \textsubscript{±.01} & 0.2554 \textsubscript{±.01} & 0.3192 \textsubscript{±.04} & 0.2303 \textsubscript{±.11} & 0.2505 \textsubscript{±.01} \\
\midrule
GraphCL & 0.2522 \textsubscript{±.06} & 0.2404 \textsubscript{±.05} & 0.2823 \textsubscript{±.02} & 0.2570 \textsubscript{±.03} & 0.3616 \textsubscript{±.07} & 0.2143 \textsubscript{±.04} & 0.5117 \textsubscript{±.07} & 0.4947 \textsubscript{±.07} & 0.4622 \textsubscript{±.09} & 0.4441 \textsubscript{±.09} \\
SimGRACE & 0.3383 \textsubscript{±.08} & 0.3294 \textsubscript{±.08} & 0.2973 \textsubscript{±.03} & 0.2869 \textsubscript{±.04} & 0.3658 \textsubscript{±.10} & 0.2619 \textsubscript{±.13} & 0.5085 \textsubscript{±.12} & 0.4882 \textsubscript{±.09} & \uline{0.5004} \textsubscript{±.03} & \uline{0.4511} \textsubscript{±.02} \\
GPF & 0.2637 \textsubscript{±.09} & 0.1887 \textsubscript{±.09} & 0.2642 \textsubscript{±.03} & 0.2169 \textsubscript{±.03} & 0.3465 \textsubscript{±.07} & 0.1820 \textsubscript{±.04} & 0.4472 \textsubscript{±.06} & 0.4463 \textsubscript{±.07} & 0.3306 \textsubscript{±.11} & 0.3287 \textsubscript{±.10} \\
GraphPrompt & 0.2798 \textsubscript{±.05} & 0.2869 \textsubscript{±.04} & 0.3794 \textsubscript{±.09} & 0.3340 \textsubscript{±.10} & 0.3530 \textsubscript{±.06} & 0.3191 \textsubscript{±.05} & 0.4967 \textsubscript{±.05} & 0.4507 \textsubscript{±.09} & 0.3703 \textsubscript{±.09} & 0.3611 \textsubscript{±.05} \\
\midrule
GCOPE & 0.3112 \textsubscript{±.06} & 0.2683 \textsubscript{±.08} & 0.3823 \textsubscript{±.10} & \textbf{0.3712} \textsubscript{±.09} & 0.3965 \textsubscript{±.06} & 0.3609 \textsubscript{±.06} & 0.4635 \textsubscript{±.08} & 0.5604 \textsubscript{±.09} & 0.4540 \textsubscript{±.06} & 0.3656 \textsubscript{±.06} \\
SAMGPT & \uline{0.4495} \textsubscript{±.09} & \uline{0.3299} \textsubscript{±.09} & \uline{0.3955} \textsubscript{±.10} & 0.3551 \textsubscript{±.09} & \uline{0.4480} \textsubscript{±.08} & \uline{0.3821} \textsubscript{±.06} & \uline{0.5671} \textsubscript{±.10} & \uline{0.5701} \textsubscript{±.10} & 0.4703 \textsubscript{±.07} & 0.4430 \textsubscript{±.06} \\
MDGCL & \textbf{0.4666} \textsubscript{±.01} & \textbf{0.4213} \textsubscript{±.05} & \textbf{0.4112} \textsubscript{±.03} & \uline{0.3575} \textsubscript{±.08} & \textbf{0.5188} \textsubscript{±.08} & \textbf{0.3839} \textsubscript{±.06} & \textbf{0.6767} \textsubscript{±.04} & \textbf{0.6415} \textsubscript{±.04} & \textbf{0.5434} \textsubscript{±.07} & \textbf{0.4849} \textsubscript{±.07} \\
\bottomrule
\end{tabular}
}
\end{table}
}

{\large
\renewcommand{\arraystretch}{1.5}
\begin{table}[t]
\centering
\caption{Accuracy(mean±std) of one-shot graph classification on homophilic and heterophilic graphs.
The best method in each column is bolded.}
\label{tab:Homophily}
\resizebox{1.0\textwidth}{!}{
\begin{tabular}{llccc ccc ccc}
\toprule
\multirow{2}{*}{Method} & \multicolumn{3}{c}{Homophilic Graphs} & \multicolumn{3}{c}{Heterophilic Graphs} \\
\cmidrule(r){2-4} \cmidrule(r){5-7}
& Cora & PubMed & Photo & Squirrel & Chameleon & Amazon-ratings \\
\midrule
GCOPE & 0.4444 \textsubscript{±.01} & 0.4815 \textsubscript{±.01} & 0.5846 \textsubscript{±.04} & 0.2505 \textsubscript{±.01} & 0.3039 \textsubscript{±.02} & 0.2442 \textsubscript{±.06} \\
SAMGPT & 0.5028 \textsubscript{±.01} & 0.5062 \textsubscript{±.10} & 0.5987 \textsubscript{±.09} & 0.1963 \textsubscript{±.03} & 0.2287 \textsubscript{±.04} & 0.1958 \textsubscript{±.07} \\
MDGCL & \textbf{0.5111} \textsubscript{±.03} & \textbf{0.6138} \textsubscript{±.04} & \textbf{0.6018} \textsubscript{±.02} & \textbf{0.2608} \textsubscript{±.02} & \textbf{0.3142} \textsubscript{±.03} & \textbf{0.2883} \textsubscript{±.01} \\
\bottomrule
\end{tabular}
}
\end{table}
}

\subsection{Performance evaluation (RQ1)}
We compare MDGCL with eight baseline methods on one-shot node and graph classification tasks.
We sample five random seeds, reporting the average results and standard deviations in Table \ref{tab:Graph_2_metric} and Table \ref{tab:Node_2_metric}, where each column represents a target domain, using other columns as source domains. 

Based on experimental results, we make the following observations:
(1) MDGCL demonstrates superior performance across both tasks, indicating that capturing domain differences and transferring domain-level knowledge significantly enhances the generalization ability of pre-trained graph models. 
(2) The single-domain pre-training methods fail to maintain consistent superiority over end-to-end methods in all downstream tasks, demonstrating that single-domain pre-training strategies cannot work effectively in multi-domain scenarios.
(3) MDGCL surpasses the other two multi-domain pre-training methods: GCOPE and SAMGPT, which mainly focus on structure and feature alignment but still adopt contrastive pre-training strategies designed for single-domain pre-training.
This demonstrates that capturing domain-specific differences is more important than aligning features and structures in multi-domain scenarios.
Additionally, we also evaluate our proposed MDGCL on 3-shot and 5-shot settings, shown in the Appendix.
\subsection{Homophily analysis (RQ2)}
Prior works\cite{Homophily_1, Homophily_2} have suggested that identical GNN architectures may exhibit substantial performance gaps when applied to homophilic and heterophilic graphs.
To evaluate the robustness of MDGCL across homophilic and heterophilic graphs, we conduct one-shot graph classification experiments on three homophilic graphs (Cora, PubMed, Photo) and three heterophilic graphs (Squirrel\cite{Squi_Chame}, Chameleon\cite{Squi_Chame}, Amazon-ratings\cite{Amazon_ratings}). 
Following the same settings as before, each dataset is treated as the target domain in turn, while the remaining five serve as source domains for pre-training. 
We compare MDGCL with two multi-domain pre-training methods: GCOPE and SAMGPT, with results reported in Table \ref{tab:Homophily}.

From the results, we make the following observations:
(1) MDGCL consistently outperforms GCOPE and SAMGPT on both homophilic and heterophilic graphs, suggesting that our method is more effective at bridging the distributional gap between them.
(2) The incorporation of heterophilic graphs introduces greater cross-domain discrepancies, while the inferior performance of GCOPE and SAMGPT confirms their tendency to project all domains into similar latent distributions, as shown in Section \ref{sec:2}.  
This uniform mapping proves particularly detrimental when handling domains with substantial domain-specific discrepancies.

\subsection{Model ablation (RQ3)}
To validate the impact of each component on the model's performance, we conduct experiments on various MDGCL variants. 
Here, MDGCL-V1 indicates our model excluding the domain tokens and the attention mechanism. 
MDGCL-V2 denotes our model excluding the domain tokens in pre-training stage.
MDGCL-V3 denotes that our model excludes the attention mechanism in the downstream stage.
The accuracy(mean±std) of MDGCL and three variants across different tasks and target domains is presented in Table \ref{tab:Model_Abla}.

The experimental results lead to following observations:
(1) MDGCL-V1 performs worse than MDGCL-V2 and MDGCL-V3. 
This demonstrates that each component contributes to enhancing the model's generalization capability. 
(2) The complete MDGCL outperforms MDGCL-V2 and MDGCL-V3, indicating that domain tokens and the attention mechanism work synergistically rather than in conflict.

\begin{wrapfigure}{r}{0.38\textwidth} 
\centering
\includegraphics[width=0.38\textwidth, height=6.5cm]{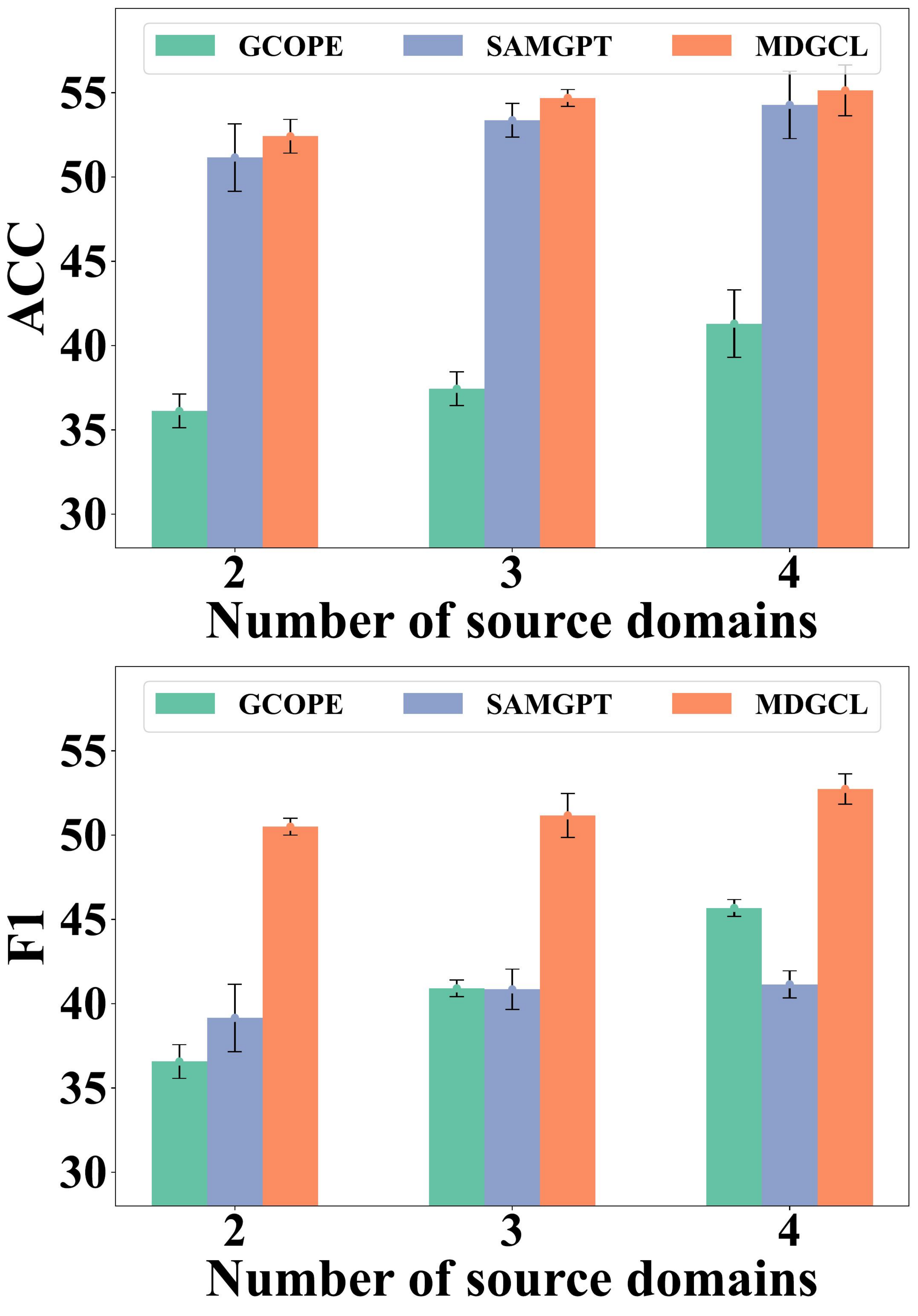} 
\caption{Data ablation study with different numbers of source domains, fixing Cora as the target domain.}
\label{fig:Data_Ablation}
\end{wrapfigure}

\subsection{Data ablation (RQ4)}
A prominent feature of foundation models is the ability to absorb knowledge from more different data domains and obtain better downstream performance.
Thus, in this section, we investigate the impact of the source domain numbers on the downstream target domain performance of our method and other text-free graph foundation models, including SAMGPT and GCOPE.

We conduct the one-shot graph classification experiment, and the result is shown in Figure \ref{fig:Data_Ablation}. 
As regard the number of source domains, 2 represents using CiteSeer and PubMed as source domains, 3 represents incorporating Photo into source domains, and 4 represents incorporating Photo and Computers. 

The experimental results yield the following observations: (1) When the number of source domains is severely limited, MDGCL still achieves competitive performance, which demonstrates that our proposed pre-training and transferring method is more suitable for multi-domain scenarios. (2) MDGCL shows improved downstream performance as the number of source domains increases, indicating that our method has the potential for pre-training on large-scale data and can serve as a pioneer for text-free graph foundation models.

{\large
\renewcommand{\arraystretch}{1.5}
\begin{table}[t]
\centering
\caption{Model ablation study on components of MDGCL.}
\label{tab:Model_Abla}
\resizebox{1.0\textwidth}{!}{
\begin{tabular}{ccc ccc ccc}
\toprule
\multirow{2}{*}{Method} & \multirow{2}{*}{\shortstack{Domain\\Tokens}} & \multirow{2}{*}{\shortstack{Attention\\Mechanism}} & \multicolumn{3}{c}{Target domain for graph classification}  & \multicolumn{3}{c}{Target domain for node classification} \\
\cmidrule(r){4-6} \cmidrule(r){7-9} & & & Cora & Photo & PubMed & Cora & Photo & PubMed \\
\midrule
MDGCL-V1 & $\times$ & $\times$ & 0.4807 \textsubscript{±.06} & 0.5867 \textsubscript{±.07} & 0.5041 \textsubscript{±.04} & 0.4017 \textsubscript{±.08} & 0.5838 \textsubscript{±.09} & 0.4603 \textsubscript{±.04} \\
MDGCL-V2 & $\times$ & $\checkmark$ & 0.5333 \textsubscript{±.06} & 0.6136 \textsubscript{±.03} & 0.5312 \textsubscript{±.08} & 0.4135 \textsubscript{±.07} & 0.5734 \textsubscript{±.09} & 0.4763 \textsubscript{±.07} \\
MDGCL-V3 & $\checkmark$ & $\times$ & 0.5252 \textsubscript{±.05} & 0.6358 \textsubscript{±.02} & 0.5600 \textsubscript{±.02} & 0.3980 \textsubscript{±.07} & 0.5942 \textsubscript{±.09} & 0.4745 \textsubscript{±.03} \\
MDGCL & $\checkmark$ & $\checkmark$ & \textbf{0.5514} \textsubscript{±.02} & \textbf{0.6617} \textsubscript{±.05} & \textbf{0.5608} \textsubscript{±.02} & \textbf{0.4666} \textsubscript{±.09} & \textbf{0.6767} \textsubscript{±.05} & \textbf{0.5188} \textsubscript{±.06} \\
\bottomrule
\end{tabular}
}
\end{table}

}

\section{Conclusion}
\label{sec:conslusion}
In this paper, we rethink current text-free graph foundation models and propose a novel framework for multi-domain pre-training and cross-domain transfer, called MDGCL. 
As far as we know, we are the first to design the contrastive learning strategy for multi-domain pre-training scenarios, establishing a new paradigm for text-free graph foundation models.
In the pre-training stage, we devise a multi-domain graph contrastive learning strategy to recognize and capture domain differences. We also introduce domain tokens to help the GNN model encode domain-level global information in the pre-training and a domain-level attention mechanism to enable fine-grained knowledge transfer in the fine-tuning.
\paragraph{Limitation and broader impacts} Currently, we focus on building graph foundation models on homogeneous graphs. A promising future direction is to explore constructing graph foundation models on heterogeneous graphs with various node or edge types. 
We believe that our proposed MDGCL has the potential to build AI systems that can be applied in real-world applications involving text-free graphs, such as social networks, molecular graphs, and recommendation systems.

\newpage

\appendix

\section{Related Work}
\label{sec:related work}
This section provides an overview of current graph foundation models.

\paragraph{Text-attributed graph foundation models}
Inspired by foundation models in natural language processing (NLP)\cite{llama,gpt4}, a few works\cite{OneforAll,HiGPT,LMCH,GFM,KG_FM,GFM_1} have attempted to leverage textual attributes as bridges on graphs, utilizing the powerful encoding capabilities of LLMs to integrate multi-domain knowledge and enable cross-domain transfer.
OFA \cite{OneforAll} introduces nodes-of-interest to standardize different tasks with a single task representation.
HiGPT \cite{HiGPT}  introduces an in-context heterogeneous graph
tokenizer to handle distribution shifts in heterogeneity.
LMCH \cite{LMCH} designs a metapath-based corpus construction method to unify heterogeneous graph representations as languages.
However, these approaches heavily rely on textual features, preventing them from being extended to the text-free graphs which are widely prevalent.
Additionally, the substantial computational overhead of LLMs makes these methods difficult to scale efficiently.

\paragraph{Text-free graph foundation models}
Some recent works\cite{GCOPE,SAMGPT} have explored multi-domain pre-training on text-free graphs. GCOPE\cite{GCOPE} introduces virtual coordinators to connect graphs from different domains, thereby aligning semantic information and structural patterns. On the other hand, SAMGPT\cite{SAMGPT} inserts structural tokens into every layer of the GNN to align multi-domain structural patterns while transferring structural knowledge from the source domains to downstream tasks.
However, these works still rely on traditional contrastive pre-training strategies which are designed for single-domain scenarios. They overlook the domain information when constructing positive and negative pairs, preventing them from capturing the differences among domains.

\section{Further Descriptions of Datasets}

More detailed information about datasets is as follows:

\begin{table}[t]
  \centering
  \caption{Statistics of datasets.}
  \label{tab:dataset-stats}
  \begin{tabular}{lrrrr}
    \toprule
    Dataset & \#Nodes & \#Edges & \#Features & \#Labels \\
    \midrule
    Cora     & 2,708     & 5,429       & 1,433      & 7  \\
    CiteSeer & 3,327     & 9,104       & 3,703      & 6  \\
    PubMed   & 19,717    & 88,648      & 500        & 3  \\
    Photo   & 7,650      & 238,162     & 745        & 8 \\
    Computers   & 13,752    & 491,722     & 767        & 10 \\
    Chameleon & 2277     & 36101       & 2325       & 5   \\
    Squirrel  & 5201     & 217073      & 2089       & 5   \\
    Amazon-ratings & 24492 & 93050     & 300        & 5   \\

    \bottomrule
  \end{tabular}
\end{table}

\begin{itemize}
\item \textbf{Citation network: }
The Cora\cite{Cora_Citeseer} and CiteSeer\cite{Cora_Citeseer} datasets include a broad range of academic publications in computer science, with each node represented by a bag-of-words feature vector derived from paper content, along with a categorical label denoting the research topic. Similarly, the PubMed\cite{Pubmed} dataset contains scientific articles focused on diabetes, sourced from the PubMed database. Here, nodes are described by TF/IDF-weighted term frequency vectors as attributes, and each publication is labeled according to the specific diabetes type discussed in the publication.
\item \textbf{Amazon network: }
The Computers\cite{computers} and Photo\cite{Photo} datasets are constructed from Amazon co-purchase networks, where nodes correspond to products, and edges reflect frequent co-purchases between two products. Each product is characterized by a bag-of-words representation derived from customer reviews, along with a categorical label indicating its product type.
The Amazon-ratings\cite{Amazon_ratings} dataset is based on the Amazon product co-purchase network metadata dataset from SNAP Datasets, where nodes denote products, and edges connect products that are frequently bought together.
\item \textbf{Wikipedia network: }
The Chameleon\cite{Squi_Chame} and Squirrel\cite{Squi_Chame} datasets are page-page networks derived from Wikipedia, centered around particular topics. Here, nodes correspond to individual web pages, with edges representing hyperlinks between them. Each node's attributes consist of meaningful nouns extracted from the page's text. Additionally, nodes are classified based on their average monthly page traffic.

\end{itemize}

We present the statistics of all related datasets in Table \ref{tab:dataset-stats}.

{\large
\renewcommand{\arraystretch}{1.5}
\begin{table}[t]
\centering
\caption{3-shot graph classification performance (mean±std Acc/F1).
The best method in each column is bolded, and the runner-up is underlined.
}
\label{tab:Graph_3_shot}
\resizebox{1.0\textwidth}{!}{
\begin{tabular}{c cc cc cc cc cc}
\toprule
\multirow{2}{*}{Method} & \multicolumn{2}{c}{Cora} & \multicolumn{2}{c}{CiteSeer} & \multicolumn{2}{c}{PubMed} & \multicolumn{2}{c}{Photo} & \multicolumn{2}{c}{Computers} \\
\cmidrule(r){2-3} \cmidrule(r){4-5} \cmidrule(r){6-7} \cmidrule(r){8-9} \cmidrule(r){10-11}
& Acc & F1 & Acc & F1 & Acc & F1 & Acc & F1 & Acc & F1 \\
\midrule
GCN & 0.4133 \textsubscript{±.05} & 0.4255 \textsubscript{±.03} & 0.4227 \textsubscript{±.02} & 0.4338 \textsubscript{±.05} & 0.5277 \textsubscript{±.02} & 0.5545 \textsubscript{±.06} & 0.5661 \textsubscript{±.02} & 0.5177 \textsubscript{±.02} & 0.4387 \textsubscript{±.10} & 0.4291 \textsubscript{±.06} \\

GAT & 0.4571 \textsubscript{±.03} & 0.4380 \textsubscript{±.07} & 0.4775 \textsubscript{±.01} & 0.4502 \textsubscript{±.04} & 0.6140 \textsubscript{±.04} & 0.5712 \textsubscript{±.09} & 0.5749 \textsubscript{±.05} & 0.5604 \textsubscript{±.04} & 0.4071 \textsubscript{±.06} & 0.3695 \textsubscript{±.02} \\

\midrule
GraphCL 
& 0.4171 \textsubscript{±.06} & 0.4196 \textsubscript{±.06} & 0.4700 \textsubscript{±.05} & 0.4571 \textsubscript{±.06} & 0.4587 \textsubscript{±.01} & 0.3946 \textsubscript{±.01} & 0.6279 \textsubscript{±.05} & 0.5904 \textsubscript{±.05} & 0.3413 \textsubscript{±.02} & 0.3177 \textsubscript{±.03} \\
SimGRACE 
& 0.3133 \textsubscript{±.03} & 0.3261 \textsubscript{±.08} & 0.4594 \textsubscript{±.09} & 0.4450 \textsubscript{±.08} & 0.5743 \textsubscript{±.03} & 0.5438 \textsubscript{±.03} & 0.6095 \textsubscript{±.06} & 0.5813 \textsubscript{±.05} & 0.3895 \textsubscript{±.05} & 0.4208 \textsubscript{±.06} \\
GPF 
& 0.4475 \textsubscript{±.01} & 0.4469 \textsubscript{±.10} & 0.4471 \textsubscript{±.06} & 0.4342 \textsubscript{±.07} & 0.4312 \textsubscript{±.04} & 0.2912 \textsubscript{±.10} & 0.6248 \textsubscript{±.05} & 0.6025 \textsubscript{±.05} & 0.4287 \textsubscript{±.10} & 0.4328 \textsubscript{±.09} \\
GraphPrompt 
& 0.4141 \textsubscript{±.03} & 0.3988 \textsubscript{±.06} & 0.5442 \textsubscript{±.02} & 0.5021 \textsubscript{±.02} & 0.4509 \textsubscript{±.08} & 0.4358 \textsubscript{±.07} & 0.6444 \textsubscript{±.04} & 0.6342 \textsubscript{±.03} & 0.5109 \textsubscript{±.02} & 0.5448 \textsubscript{±.10} \\
\midrule

GCOPE 
& 0.6019 \textsubscript{±.03} & \uline{0.5860} \textsubscript{±.03} & \uline{0.5999} \textsubscript{±.02} & \uline{0.5649} \textsubscript{±.02} & \uline{0.6228} \textsubscript{±.03} & \uline{0.5997} \textsubscript{±.04} & 0.6742 \textsubscript{±.02} & 0.6363 \textsubscript{±.01} & \uline{0.5742} \textsubscript{±.03} & 0.5340 \textsubscript{±.04} \\

SAMGPT 
& \uline{0.6285} \textsubscript{±.09} & 0.5316 \textsubscript{±.07} & 0.5593 \textsubscript{±.06} & 0.5092 \textsubscript{±.06} & 0.5665 \textsubscript{±.07} & 0.5646 \textsubscript{±.09} & \uline{0.6804} \textsubscript{±.07} & \uline{0.6599} \textsubscript{±.06} & \textbf{0.5911} \textsubscript{±.07} & \textbf{0.5864} \textsubscript{±.09} \\
MDGCL & \textbf{0.6350} \textsubscript{±.04} & \textbf{0.6179} \textsubscript{±.04} & \textbf{0.6882} \textsubscript{±.03} & \textbf{0.6457} \textsubscript{±.03} & \textbf{0.6551} \textsubscript{±.01} & \textbf{0.6515} \textsubscript{±.01} & \textbf{0.7092} \textsubscript{±.03} & \textbf{0.6817} \textsubscript{±.02} & 0.5631 \textsubscript{±.08} & \uline{0.5496} \textsubscript{±.07} \\
\bottomrule
\end{tabular}
}
\end{table}
}

{\large
\renewcommand{\arraystretch}{1.5}
\begin{table}[t]
\centering
\caption{3-shot node classification performance (mean±std Acc/F1).
The best method in each column is bolded, and the runner-up is underlined.}
\label{tab:Node_3_shot}
\resizebox{1.0\textwidth}{!}{
\begin{tabular}{c cc cc cc cc cc}
\toprule
\multirow{2}{*}{Method} & \multicolumn{2}{c}{Cora} & \multicolumn{2}{c}{CiteSeer} & \multicolumn{2}{c}{PubMed} & \multicolumn{2}{c}{Photo} & \multicolumn{2}{c}{Computers} \\
\cmidrule(r){2-3} \cmidrule(r){4-5} \cmidrule(r){6-7} \cmidrule(r){8-9} \cmidrule(r){10-11}
& Acc & F1 & Acc & F1 & Acc & F1 & Acc & F1 & Acc & F1 \\
\midrule
GCN 
& 0.3955 \textsubscript{±.02} & 0.4109 \textsubscript{±.10} & 0.4107 \textsubscript{±.01} & 0.3995 \textsubscript{±.08} & 0.5262 \textsubscript{±.05} & 0.5337 \textsubscript{±.01} & 0.5849 \textsubscript{±.02} & 0.5930 \textsubscript{±.03} & 0.4818 \textsubscript{±.06} & 0.4710 \textsubscript{±.05} \\
GAT 
& 0.4417 \textsubscript{±.01} & 0.4185 \textsubscript{±.03} & 0.4509 \textsubscript{±.02} & 0.4289 \textsubscript{±.04} & \uline{0.6015} \textsubscript{±.07} & \uline{0.6089} \textsubscript{±.06} & 0.5773 \textsubscript{±.02} & 0.5630 \textsubscript{±.06} & 0.4223 \textsubscript{±.10} & 0.4073 \textsubscript{±.09} \\
\midrule
GraphCL 
& 0.3595 \textsubscript{±.06} & 0.3565 \textsubscript{±.04} & 0.3108 \textsubscript{±.06} & 0.2956 \textsubscript{±.06} & 0.4421 \textsubscript{±.07} & 0.3411 \textsubscript{±.14} & 0.6115 \textsubscript{±.13} & 0.6182 \textsubscript{±.11} & \uline{0.6150} \textsubscript{±.02} & 0.6007 \textsubscript{±.03} \\
SimGRACE 
& 0.4551 \textsubscript{±.04} & 0.4526 \textsubscript{±.03} & 0.3404 \textsubscript{±.03} & 0.3296 \textsubscript{±.03} & 0.4349 \textsubscript{±.07} & 0.3666 \textsubscript{±.12} & 0.6218 \textsubscript{±.09} & 0.6044 \textsubscript{±.10} & 0.5243 \textsubscript{±.09} & 0.5300 \textsubscript{±.08} \\
GPF 
& 0.2997 \textsubscript{±.01} & 0.2823 \textsubscript{±.07} & 0.2539 \textsubscript{±.05} & 0.2881 \textsubscript{±.10} & 0.3982 \textsubscript{±.01} & 0.2883 \textsubscript{±.03} & 0.4566 \textsubscript{±.08} & 0.4450 \textsubscript{±.06} & 0.3903 \textsubscript{±.02} & 0.3842 \textsubscript{±.10} \\
GraphPrompt 
& 0.3010 \textsubscript{±.06} & 0.3359 \textsubscript{±.05} & 0.3521 \textsubscript{±.10} & 0.2997 \textsubscript{±.09} & 0.4291 \textsubscript{±.07} & 0.3882 \textsubscript{±.03} & 0.5010 \textsubscript{±.08} & 0.4507 \textsubscript{±.05} & 0.5114 \textsubscript{±.09} & 0.4518 \textsubscript{±.04} \\
\midrule
GCOPE 
& 0.4819 \textsubscript{±.07} & 0.4684 \textsubscript{±.07} & \uline{0.5468} \textsubscript{±.06} & \textbf{0.5096} \textsubscript{±.05} & 0.4901 \textsubscript{±.06} & 0.4106 \textsubscript{±.09} & 0.6978 \textsubscript{±.08} & 0.7055 \textsubscript{±.09} & 0.5737 \textsubscript{±.02} & \uline{0.6023} \textsubscript{±.01} \\
SAMGPT 
& \uline{0.6180} \textsubscript{±.07} & \uline{0.4783} \textsubscript{±.05} & \textbf{0.5532} \textsubscript{±.10} & \uline{0.5093} \textsubscript{±.04} & 0.4883 \textsubscript{±.04} & 0.4563 \textsubscript{±.05} & \uline{0.7122} \textsubscript{±.10} & \uline{0.7169} \textsubscript{±.08} & 0.6047 \textsubscript{±.05} & 0.5840 \textsubscript{±.07} \\
MDGCL 
& \textbf{0.6346} \textsubscript{±.05} & \textbf{0.6232} \textsubscript{±.04} & 0.5326 \textsubscript{±.02} & 0.5027 \textsubscript{±.01} & \textbf{0.6392} \textsubscript{±.07} & \textbf{0.6343} \textsubscript{±.07} & \textbf{0.7563} \textsubscript{±.04} & \textbf{0.7454} \textsubscript{±.03} & \textbf{0.6199} \textsubscript{±.09} & \textbf{0.6140} \textsubscript{±.10} \\
\bottomrule
\end{tabular}
}
\end{table}
}

{\large
\renewcommand{\arraystretch}{1.5}
\begin{table}[t]
\centering
\caption{5-shot graph classification performance (mean±std Acc/F1).
The best method in each column is bolded, and the runner-up is underlined.
}
\label{tab:Graph_5_shot}
\resizebox{1.0\textwidth}{!}{
\begin{tabular}{c cc cc cc cc cc}
\toprule
\multirow{2}{*}{Method} & \multicolumn{2}{c}{Cora} & \multicolumn{2}{c}{CiteSeer} & \multicolumn{2}{c}{PubMed} & \multicolumn{2}{c}{Photo} & \multicolumn{2}{c}{Computers} \\
\cmidrule(r){2-3} \cmidrule(r){4-5} \cmidrule(r){6-7} \cmidrule(r){8-9} \cmidrule(r){10-11}
& Acc & F1 & Acc & F1 & Acc & F1 & Acc & F1 & Acc & F1 \\
\midrule
GCN & 
0.6017 \textsubscript{±.03} & 0.5982 \textsubscript{±.06} & 0.4617 \textsubscript{±.04} & 0.4725 \textsubscript{±.05} & 0.6049 \textsubscript{±.01} & 0.5770 \textsubscript{±.03} & 0.5998 \textsubscript{±.04} & 0.6090 \textsubscript{±.06} & 0.4275 \textsubscript{±.04} & 0.4160 \textsubscript{±.02} \\

GAT & 
0.5662 \textsubscript{±.01} & 0.5733 \textsubscript{±.04} & 0.5091 \textsubscript{±.02} & 0.4936 \textsubscript{±.05} & 0.5982 \textsubscript{±.01} & 0.6102 \textsubscript{±.02} & 0.6188 \textsubscript{±.04} & 0.5908 \textsubscript{±.03} & 0.5146 \textsubscript{±.09} & 0.5203 \textsubscript{±.08} \\

\midrule
GraphCL 
& 0.5431 \textsubscript{±.07} & 0.5562 \textsubscript{±.05} & 0.5306 \textsubscript{±.04} & 0.5269 \textsubscript{±.02} & 0.5228 \textsubscript{±.02} & 0.4771 \textsubscript{±.02} & 0.6358 \textsubscript{±.03} & 0.6076 \textsubscript{±.03} & 0.3112 \textsubscript{±.03} & 0.3128 \textsubscript{±.03} \\

SimGRACE 
& 0.4852 \textsubscript{±.02} & 0.4889 \textsubscript{±.04} & 0.5432 \textsubscript{±.07} & 0.5333 \textsubscript{±.05} & \uline{0.6288} \textsubscript{±.04} & 0.6015 \textsubscript{±.05} & 0.6484 \textsubscript{±.06} & 0.6234 \textsubscript{±.05} & 0.4158 \textsubscript{±.01} & 0.4392 \textsubscript{±.02} \\

GPF 
& 0.5306 \textsubscript{±.09} & 0.5398 \textsubscript{±.04} & 0.5295 \textsubscript{±.04} & 0.5209 \textsubscript{±.04} & 0.5392 \textsubscript{±.11} & 0.4503 \textsubscript{±.12} & 0.6439 \textsubscript{±.02} & 0.6171 \textsubscript{±.03} & 0.4496 \textsubscript{±.06} & 0.4466 \textsubscript{±.05} \\

GraphPrompt 
& 0.4544 \textsubscript{±.04} & 0.4259 \textsubscript{±.05} & 0.4816 \textsubscript{±.07} & 0.4290 \textsubscript{±.08} & 0.5510 \textsubscript{±.02} & 0.5394 \textsubscript{±.03} & 0.6751 \textsubscript{±.04} & 0.6745 \textsubscript{±.01} & 0.5008 \textsubscript{±.01} & 0.4919 \textsubscript{±.03} \\
\midrule

GCOPE 
& 0.6877 \textsubscript{±.01} & \uline{0.6812} \textsubscript{±.01} & \uline{0.6940} \textsubscript{±.01} & \uline{0.6719} \textsubscript{±.01} & 0.6209 \textsubscript{±.01} & \uline{0.6165} \textsubscript{±.01} & 0.6979 \textsubscript{±.02} & 0.6281 \textsubscript{±.02} & 0.5502 \textsubscript{±.01} & 0.5393 \textsubscript{±.03} \\

SAMGPT 
& \textbf{0.7385} \textsubscript{±.07} & 0.5586 \textsubscript{±.06} & 0.6022 \textsubscript{±.05} & 0.5496 \textsubscript{±.05} & 0.5891 \textsubscript{±.07} & 0.5928 \textsubscript{±.07} & \uline{0.7069} \textsubscript{±.06} & \textbf{0.7004} \textsubscript{±.06} & \uline{0.6018} \textsubscript{±.07} & \textbf{0.6117} \textsubscript{±.06} \\

MDGCL &
\uline{0.7075} \textsubscript{±.02} & \textbf{0.7059} \textsubscript{±.01} & \textbf{0.7225} \textsubscript{±.02} & \textbf{0.6995} \textsubscript{±.02} & \textbf{0.6501} \textsubscript{±.02} & \textbf{0.6522} \textsubscript{±.02} & \textbf{0.7152} \textsubscript{±.03} & \uline{0.6998} \textsubscript{±.02} & \textbf{0.6263} \textsubscript{±.01} & \uline{0.5918} \textsubscript{±.03} \\
\bottomrule
\end{tabular}
}
\end{table}
}

{\large
\renewcommand{\arraystretch}{1.5}
\begin{table}[t]
\centering
\caption{5-shot node classification performance (mean±std Acc/F1).
The best method in each column is bolded, and the runner-up is underlined.}
\label{tab:Node_5_shot}
\resizebox{1.0\textwidth}{!}{
\begin{tabular}{c cc cc cc cc cc}
\toprule
\multirow{2}{*}{Method} & \multicolumn{2}{c}{Cora} & \multicolumn{2}{c}{CiteSeer} & \multicolumn{2}{c}{PubMed} & \multicolumn{2}{c}{Photo} & \multicolumn{2}{c}{Computers} \\
\cmidrule(r){2-3} \cmidrule(r){4-5} \cmidrule(r){6-7} \cmidrule(r){8-9} \cmidrule(r){10-11}
& Acc & F1 & Acc & F1 & Acc & F1 & Acc & F1 & Acc & F1 \\
\midrule
GCN 
& 0.5905 \textsubscript{±.07} & \uline{0.5919} \textsubscript{±.10} & 0.4217 \textsubscript{±.03} & 0.4088 \textsubscript{±.05} & 0.6009 \textsubscript{±.06} & 0.5932 \textsubscript{±.04} & 0.6099 \textsubscript{±.05} & 0.5915 \textsubscript{±.02} & 0.4557 \textsubscript{±.04} & 0.4311 \textsubscript{±.05} \\
GAT 
& 0.5866 \textsubscript{±.02} & 0.5475 \textsubscript{±.02} & 0.4961 \textsubscript{±.01} & 0.4819 \textsubscript{±.04} & \uline{0.6036} \textsubscript{±.08} & \uline{0.5979} \textsubscript{±.09} & 0.6237 \textsubscript{±.09} & 0.6358 \textsubscript{±.10} & 0.5216 \textsubscript{±.07} & 0.4773 \textsubscript{±.10} \\
\midrule

GraphCL 
& 0.4031 \textsubscript{±.11} & 0.4022 \textsubscript{±.09} & 0.3737 \textsubscript{±.02} & 0.3576 \textsubscript{±.02} & 0.4368 \textsubscript{±.05} & 0.2887 \textsubscript{±.12} & 0.6523 \textsubscript{±.08} & 0.6494 \textsubscript{±.06} & 0.6586 \textsubscript{±.02} & 0.6642 \textsubscript{±.01} \\

SimGRACE 
& 0.5162 \textsubscript{±.04} & 0.5136 \textsubscript{±.03} & 0.3814 \textsubscript{±.05} & 0.3666 \textsubscript{±.03} & 0.4788 \textsubscript{±.09} & 0.3688 \textsubscript{±.13} & 0.6841 \textsubscript{±.06} & 0.6981 \textsubscript{±.05} & 0.6118 \textsubscript{±.06} & \uline{0.6817} \textsubscript{±.05} \\

GPF 
& 0.2901 \textsubscript{±.09} & 0.2739 \textsubscript{±.08} & 0.3723 \textsubscript{±.04} & 0.3087 \textsubscript{±.05} & 0.4557 \textsubscript{±.07} & 0.3733 \textsubscript{±.03} & 0.5652 \textsubscript{±.11} & 0.5436 \textsubscript{±.13} & 0.4292 \textsubscript{±.12} & 0.4763 \textsubscript{±.15} \\

GraphPrompt 
& 0.3714 \textsubscript{±.09} & 0.3659 \textsubscript{±.07} & 0.3993 \textsubscript{±.06} & 0.3877 \textsubscript{±.08} & 0.4914 \textsubscript{±.09} & 0.4807 \textsubscript{±.13} & 0.5513 \textsubscript{±.02} & 0.4976 \textsubscript{±.03} & 0.4215 \textsubscript{±.01} & 0.4378 \textsubscript{±.01} \\
\midrule

GCOPE 
& 0.5647 \textsubscript{±.05} & 0.5640 \textsubscript{±.05} & \uline{0.5623} \textsubscript{±.04} & \uline{0.5302} \textsubscript{±.03} & 0.4610 \textsubscript{±.06} & 0.3593 \textsubscript{±.09} & \uline{0.8055} \textsubscript{±.05} & \uline{0.7836} \textsubscript{±.05} & 0.6664 \textsubscript{±.04} & 0.6687 \textsubscript{±.03} \\

SAMGPT 
& \uline{0.6665} \textsubscript{±.08} & 0.5220 \textsubscript{±.08} & \textbf{0.6053 \textsubscript{±.05}} & \textbf{0.5563 \textsubscript{±.05}} & 0.5296 \textsubscript{±.07} & 0.5115 \textsubscript{±.09} & 0.7568 \textsubscript{±.05} & 0.7491 \textsubscript{±.05} & \uline{0.6808} \textsubscript{±.07} & 0.6644 \textsubscript{±.07} \\
MDGCL 
& \textbf{0.6751 \textsubscript{±.06}} & \textbf{0.6668 \textsubscript{±.06}} & 0.5394 \textsubscript{±.02} & 0.5027 \textsubscript{±.02} & \textbf{0.6372 \textsubscript{±.04}} & \textbf{0.6300 \textsubscript{±.05}} & \textbf{0.8453 \textsubscript{±.04}} & \textbf{0.8206 \textsubscript{±.04}} & \textbf{0.7080 \textsubscript{±.04}} & \textbf{0.6833 \textsubscript{±.04}} \\
\bottomrule
\end{tabular}
}
\end{table}
}

\section{Implementation Details}
We present the key settings of all baselines and our proposed MDGCL.

\paragraph{Operating environment}
The environment where our code runs is shown as follows:

\begin{itemize}

\item Operating system: 
Linux version 6.8.0-59-generic.x86\_64.
\item CPU information:
Intel Xeon Gold 6348 CPU @ 2.60GHz.
\item GPU information: Nvidia GeForce RTX 3090.

\end{itemize}

\paragraph{Baseline settings.}
For GCN\cite{GCN}, we employ a 2-layer architecture.
For GAT\cite{GAT}, we adopt the setting in SAMGPT. It is a 2-layer architecture, with 8 attention heads in the first layer and 1 attention head in the second layer.
For GraphCL\cite{GCL}, we generate augmented view pairs by randomly selecting two distinct strategies from edge perturbation, node dropping, and feature masking. 
For all baselines, we set the input dimension to 50 and the hidden dimension to 256. 

\paragraph{MDGCL settings.}
For our proposed MDGCL, we adopt a 2-layer GCN as the backbone with the input dimension of 50 and the hidden dimension of 256, consistent with all graph pre-training methods.
During the pre-training stage, we sample 50 subgraphs per source domain via random walks with a path length of 50, and construct domain tokens through sum pooling of node features. 
In the downstream stage, we implement the domain-level attention mechanism with 2 attention heads.

\section{Few-shot Performance}
We compare our proposed MDGCL with all baseline methods on both node classification and graph classification tasks under 3-shot and 5-shot settings, with the experimental results presented in Table \ref{tab:Graph_3_shot}, Table \ref{tab:Node_3_shot}, Table \ref{tab:Graph_5_shot}, and Table \ref{tab:Node_5_shot}.
We observe that MDGCL consistently and significantly outperforms all baseline methods, demonstrating our method's capacity to universally enhance the model's generalization ability in low-shot scenarios.

Furthermore, as the number of shots increases, single-domain pre-training methods perform worse than end-to-end models. This performance gap emerges because graph pre-training techniques are primarily designed to address label scarcity.
In contrast, when sufficient supervision is available, end-to-end models can better leverage labeled data to improve performance.

\section{Complexity Analysis}
\label{sec:complexity}

\begin{table}[h]
\centering
\caption{Training Efficiency}
\label{tab:complexity}
\begin{tabular}{l S[table-format=2.2] S[table-format=5.0]}
\toprule
{Model} & {Time per Epoch (\si{\second})} & {GPU Memory (\si{\mega\byte})} \\
\midrule
GCOPE   & 56.67  & 344    \\
SAMGPT  & 0.52   & 11588  \\
MDGCL   & 0.48   & 1438   \\
\bottomrule
\end{tabular}
\end{table}

In pre-training efficiency, we compare our proposed MDGCL against two text-free graph foundation models, GCOPE and SAMGPT, with results shown in Table \ref{tab:complexity}. 
Compared to GCOPE's slower pre-training and SAMGPT's higher memory consumption, our MDGCL achieves superior efficiency in training speed and memory utilization.

During the pre-training stage with $M$ source domains, we sample K subgraphs from each source domain, where any two subgraphs from the same domain form a positive pair, resulting in a total of $M \times C_K^2$ positive pairs.
For negative pairs, we construct N cross-domain negative pairs between each two of the distinct source domains, resulting in a total of $C_M^2 \times N$ negative pairs.
This demonstrates the number of training samples depends solely on $K$, $N$, and $M$, independent of the scale of dataset. 
As a result, our method maintains computational efficiency and provable scalability even for large-scale graphs.

\section{Hyperparameter Analysis}

\subsection{Walk length}

\begin{figure}[htbp]
    \centering
    \includegraphics[width=\linewidth]{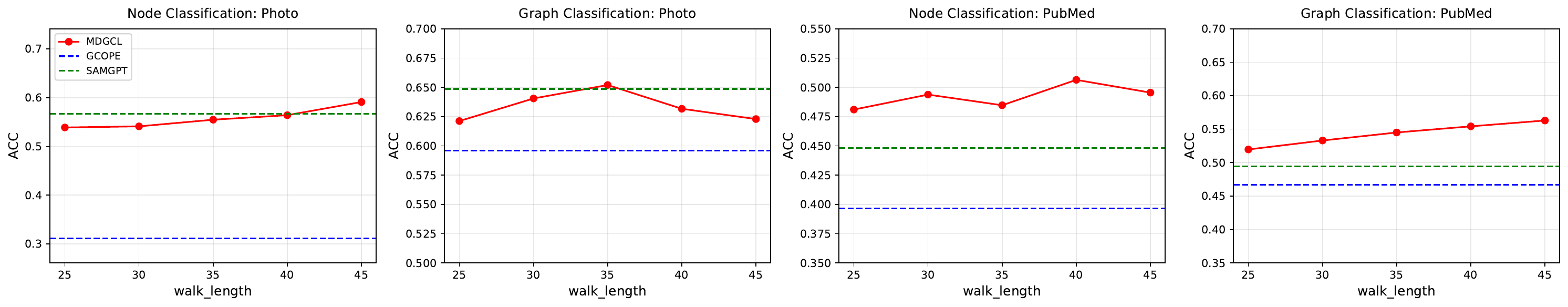}
    \caption{Analysis of the walk length.}
    \label{FIG:walk_len}
\end{figure}

We evaluate the performance of MDGCL with different walk lengths, as shown in Figure \ref{FIG:walk_len}.
From the results, we observe consistent performance improvement with increasing walk length in most cases, as longer walks capture richer domain information, thereby enhancing the GNN's ability to discern domain-specific difference.

\subsection{Number of negative pairs}

\begin{figure}[htbp]
    \centering
    \includegraphics[width=\linewidth]{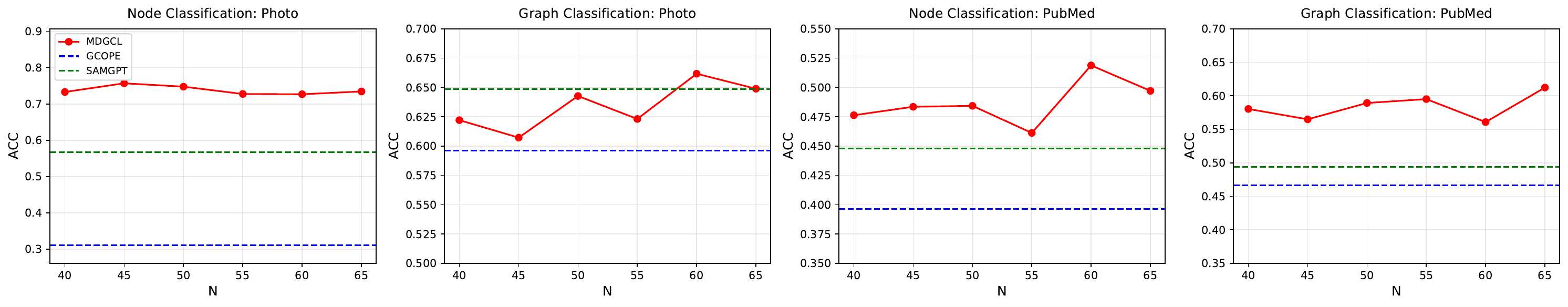}
    \caption{Analysis of the hyperparameter N.}
    \label{FIG:neg}
\end{figure}

As mentioned in Section \ref{sec:complexity}, the number of negative samples during the pre-training stage is controlled by the hyperparameter N.
We evaluate the performance of MDGCL with different numbers of negative samples, as shown in Figure \ref{FIG:neg}.
We observe that even when the number of negative samples is small, our model still achieves state-of-the-art performance in most cases,  highlighting the effectiveness of our multi-domain contrastive learning strategy in capturing domain-specific discrepancy.

\bibliographystyle{unsrt}
\bibliography{reference}

\end{document}